\documentclass[10pt,twocolumn,letterpaper]{article}

\usepackage{cvpr}
\usepackage{times}
\usepackage{epsfig}
\usepackage{graphicx}
\usepackage{amsmath}
\usepackage{amssymb}
\usepackage{multirow}
\usepackage{pifont}
\usepackage{subcaption}
\usepackage[font=small,labelfont=bf]{caption}

\usepackage[pagebackref=true,breaklinks=true,letterpaper=true,colorlinks,bookmarks=false]{hyperref}

\cvprfinalcopy 

\def\cvprPaperID{794} 

\ifcvprfinal\fi
\begin{document}

\title{Visual7W: Grounded Question Answering in Images}

\author{Yuke Zhu$^\dagger$ \qquad Oliver Groth$^\ddagger$
\qquad Michael Bernstein$^\dagger$
\qquad Li Fei-Fei$^\dagger$\\
$^\dagger$Computer Science Department, Stanford University\\
$^\ddagger$Computer Science Department, Dresden University of Technology
}

\maketitle


\begin{abstract}

We have seen great progress in basic perceptual tasks such as object recognition and detection. However, AI models still fail to match humans in high-level vision tasks due to the lack of capacities for deeper reasoning. Recently the new task of visual question answering (QA) has been proposed to evaluate a model's capacity for deep image understanding. Previous works have established a loose, global association between QA sentences and images. However, many questions and answers, in practice, relate to local regions in the images. 
We establish a semantic link between textual descriptions and image regions by object-level grounding. It enables a new type of QA with visual answers, in addition to textual answers used in previous work. We study the visual QA tasks in a grounded setting with a large collection of 7W multiple-choice QA pairs. 
Furthermore, we evaluate human performance and several baseline models on the QA tasks. Finally, we propose a novel LSTM model with spatial attention to tackle the 7W QA tasks.
\end{abstract}


\section{Introduction}
The recent development of deep learning technologies has achieved successes in many perceptual visual tasks such as object recognition, image classification and pose estimation ~\cite{karpathy2014large,krizhevsky2012imagenet,lin2014microsoft,ILSVRC15,simonyan2014very,taigman2014deepface,toshev2014deeppose}. Yet the status quo of computer vision is still far from matching human capabilities, especially when it comes to understanding an image in all its details.
Recently, visual question answering (QA) has been proposed as a proxy task for evaluating a vision system's capacity for deeper image understanding. Several QA datasets~\cite{antol2015vqa,gao2015you,malinowski2014multi,ren2015image,VisualMadlibs} have been released since last year. They contributed valuable data for training visual QA systems and introduced various tasks, from picking correct multiple-choice answers~\cite{antol2015vqa} to filling in blanks~\cite{VisualMadlibs}. 

\begin{figure}[t!]
\begin{center}
\includegraphics[width=1.\linewidth]{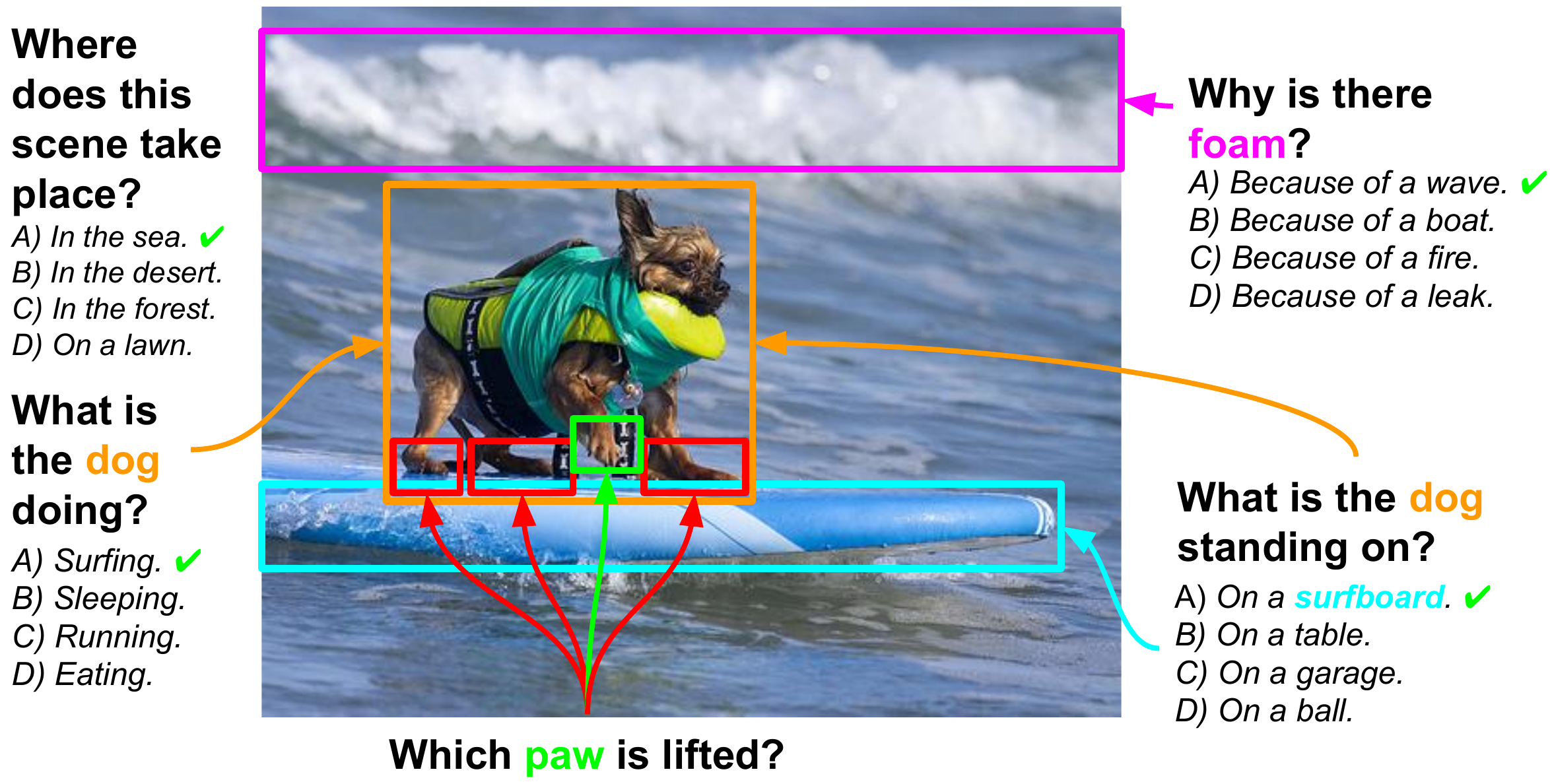}
\vspace{-7mm}
\caption{Deep image understanding relies on detailed knowledge about different image parts. We employ diverse questions to acquire detailed information on images, ground objects mentioned in text with their visual appearances, and provide a multiple-choice setting for evaluating a visual question answering task with both textual and visual answers.}
\label{fig:pull}
\vspace{-7mm}
\end{center}
\end{figure}

Pioneer work in image captioning~\cite{chen2015cvpr,Donahue_2015_CVPR,karpathy2015cvpr,Vinyals_2015_CVPR,xu2015icml}, sentence-based image retrieval~\cite{karpathy2014deep,socher2014grounded} and visual QA~\cite{antol2015vqa,gao2015you,ren2015image} shows promising results.
These works aimed at establishing a global association between sentences and images. However, as Flickr30K~\cite{plummer2015flickr30k,young2014image} and Visual Madlibs~\cite{VisualMadlibs} suggest, a tighter semantic link between textual descriptions and corresponding visual regions is a key ingredient for better models. As Fig.~\ref{fig:pull} shows, the localization of objects can be a critical step to understand images better and solve image-related questions. Providing these image-text correspondences is called \emph{grounding}. 
Inspired by Geman et al.'s prototype of a visual Turing test based on image regions~\cite{geman2015visual} and the comprehensive data collection of QA pairs on COCO images~\cite{lin2014microsoft} such as VQA~\cite{antol2015vqa} and Baidu~\cite{gao2015you}, we fuse visual QA and grounding in order to create a new QA dataset with dense annotations and a more flexible evaluation environment. Object-level grounding provides a stronger link between QA pairs and images than global image-level associations. Furthermore, it allows us to resolve coreference ambiguity~\cite{kong2014you,ramanathan2014linking} and to understand object distributions in QA, and enables visually grounded answers that consist of object bounding boxes.

Motivated by the goal of developing a model for visual QA based on grounded regions, our paper introduces a dataset that extends previous approaches~\cite{antol2015vqa,gao2015you,ren2015image} and proposes an attention-based model to perform this task.
We collected 327,939 QA pairs on 47,300 COCO images~\cite{lin2014microsoft}, together with 1,311,756 human-generated multiple-choices and 561,459 object groundings from 36,579 categories.
Our data collection was inspired by the age-old idea of the \emph{W} questions in journalism to describe a complete story~\cite{kuhn2013political}.
The \emph{7W} questions roughly correspond to an array of standard vision tasks: \emph{what}~\cite{girshick2014rich,karpathy2014large,simonyan2014very}, \emph{where}~\cite{linlearning,zhou2014learning}, \emph{when}~\cite{palermo2012dating,pickup2014seeing}, \emph{who}~\cite{ramanathan2014linking,taigman2014deepface}, \emph{why}~\cite{pirsiavash2014inferring}, \emph{how}~\cite{lampert2009learning,patterson2012sun} and \emph{which}~\cite{kazemzadeh2014emnlp,kiapour2015iccv}.
The Visual7W dataset features richer questions and longer answers than VQA~\cite{antol2015vqa}. In addition, we provide complete grounding annotations that link the object mentions in the QA sentences to their bounding boxes in the images and therefore introduce a new QA type with image regions as the visually grounded answers. We refer to questions with textual answers as \emph{telling} questions (\emph{what}, \emph{where}, \emph{when}, \emph{who}, \emph{why} and \emph{how}) and to such with visual answers as \emph{pointing} questions (\emph{which}). We provide a detailed comparison and data analysis in Sec.~\ref{sec:data_analysis}.

A salient property of our dataset is the notable gap between human performance (96.6\%) and state-of-the-art LSTM models~\cite{malinowski2015ask} (52.1\%) on the visual QA tasks. We add a new spatial attention mechanism to an LSTM architecture for tackling the visually grounded QA tasks with both textual and visual answers (see Sec.~\ref{sec:attention_model}). The model aims to capture the intuition that answers to image-related questions usually correspond with specific image regions. It learns to attend to the pertinent regions as it reads the question tokens in a sequence. 
We achieve state-of-the-art performance with 55.6\%, and find correlations between the model's attention heat maps and the object groundings (see Sec.~\ref{sec:exp}).
Due to the large performance gap between human and machine, we envision our dataset and visually grounded QA tasks to contribute to a long-term joint effort from several communities such as vision, natural language processing and knowledge to close the gap together.

The Visual7W dataset constitutes a part of the Visual Genome project~\cite{krishnavisualgenome}. Visual Genome contains 1.7 million QA pairs of the 7W question types, which offers the largest visual QA collection to date for training models. The QA pairs in Visual7W are a subset of the 1.7 million QA pairs from Visual Genome. Moreover, Visual7W includes extra annotations such as object groundings, multiple choices and human experiments, making it a clean and complete benchmark for evaluation and analysis.

\section{Related Work}
\noindent
\textbf{Vision\,+\,Language.} 
There have been years of effort in connecting the visual and textual information for joint learning~\cite{barnard2003matching,kong2014you,pirsiavash2014inferring,ramanathan2014linking,socher2014grounded,zitnick2013learning}.
Image and video captioning has become a popular task in the past year~\cite{chen2015cvpr,Donahue_2015_CVPR,karpathy2015cvpr,rohrbach2013translating,Vinyals_2015_CVPR,xu2015icml}. The goal is to generate text snippets to describe the images and regions instead of just predicting a few labels. Visual question answering is a natural extension to the captioning tasks, but is more interactive and has a stronger connection to real-world applications~\cite{bigham2010vizwiz}.

\vspace{1mm}
\noindent
\textbf{Text-based question answering.} Question answering in NLP has been a well-established problem. Successful applications can be seen in voice assistants in mobile devices, search engines and game shows (e.g., IBM Waston). Traditional question answering system relies on an elaborate pipeline of models involving natural language parsing, knowledge base querying, and answer generation~\cite{ferrucci2010}.  Recent neural network models attempt to learn end-to-end directly from questions and answers~\cite{iyyer2014emnlp,weston2015towards}. 

\vspace{1mm}
\noindent
\textbf{Visual question answering.} 
Geman et al.~\cite{geman2015visual} proposed a restricted visual Turing test to evaluate visual understanding. 
The DAQUAR dataset is the first toy-sized QA benchmark built upon indoor scene RGB-D images. Most of the other datasets~\cite{antol2015vqa,gao2015you,ren2015image,VisualMadlibs} collected QA pairs on Microsoft COCO images~\cite{lin2014microsoft}, either generated automatically by NLP tools~\cite{ren2015image} or written by human workers~\cite{antol2015vqa,gao2015you,VisualMadlibs}. Following these datasets, an array of models has been proposed to tackle the visual QA tasks. The proposed models range from probabilistic inference~\cite{malinowski2014multi,tu2014joint,zhu2014eccv} and recurrent neural networks~\cite{antol2015vqa,gao2015you,malinowski2015ask,ren2015image} to convolutional networks~\cite{ma2015cnnQA}. 
Previous visual QA datasets evaluate textual answers on images while omitting the links between the object mentions and their visual appearances. Inspired by Geman et al.~\cite{geman2015visual}, we establish the link by grounding objects in the images and perform experiments in the grounded QA setting.


\section{Creating the Visual7W Dataset}
\label{sec:7w_questions}

We elaborate on the details of the data collection we conducted upon 47,300 images from COCO~\cite{lin2014microsoft} (a subset of images from Visual Genome~\cite{krishnavisualgenome}).
We leverage the six W questions (\emph{what}, \emph{where}, \emph{when}, \emph{who}, \emph{why}, and \emph{how}) to systematically examine a model's capability for visual understanding, and append a 7th \emph{which} question category. 
This extends existing visual QA setups~\cite{antol2015vqa,gao2015you,ren2015image} to accommodate visual answers.
We standardize the visual QA tasks with multi-modal answers in a multiple-choice format. Each question comes with four answer candidates, with one being the correct answer. In addition, we ground all the objects mentioned in the QA pairs to their corresponding bounding boxes in the images. The object-level groundings enable examining the object distributions and resolve the coreference ambiguity~\cite{kong2014you,ramanathan2014linking}.


\begin{figure*}[t!]
\includegraphics[width=1.\linewidth]{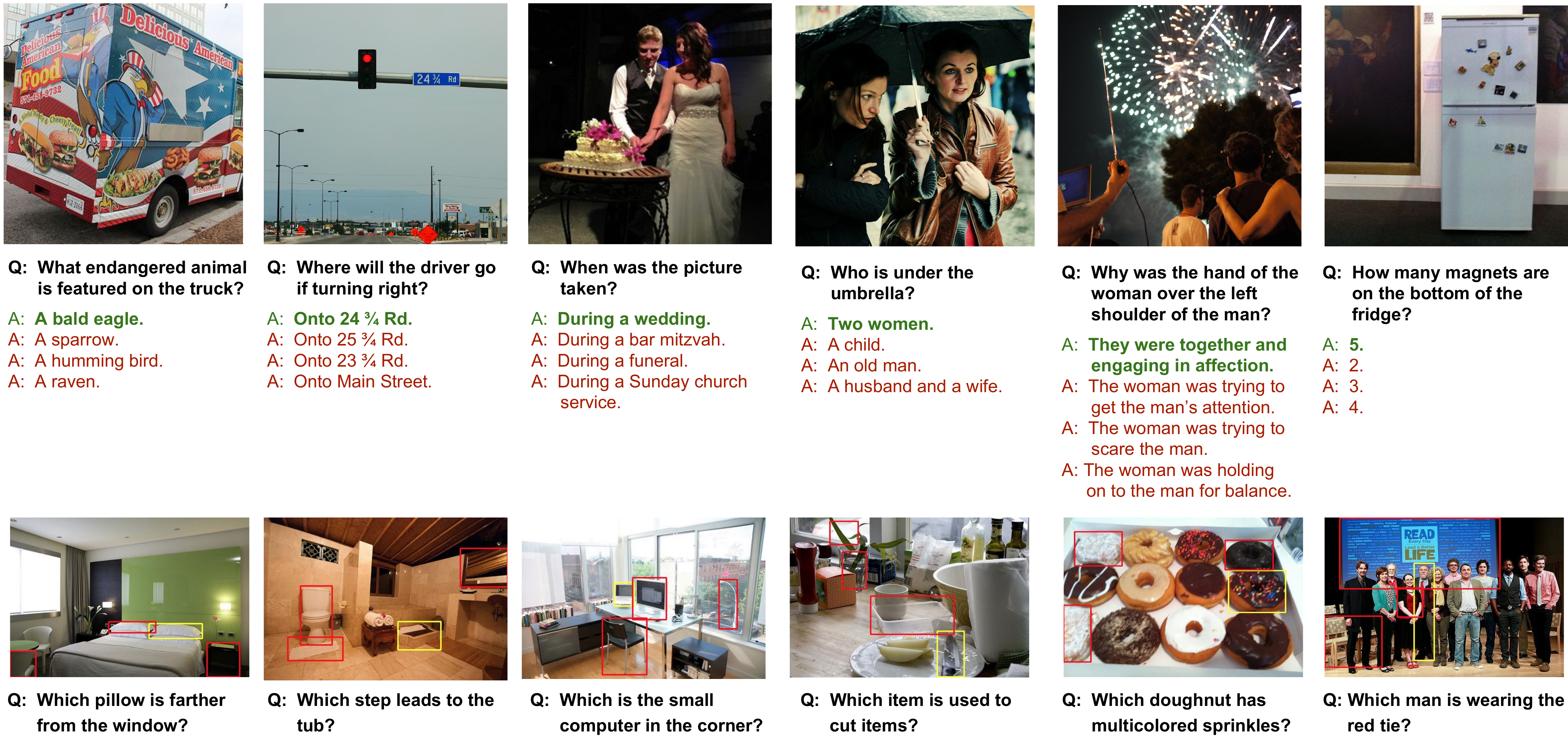}
\vspace{-6mm}
\caption{Examples of multiple-choice QA from the 7W question categories. The first row shows \emph{telling} questions where the green answer is the ground-truth, and the red ones are human-generated wrong answers. The \emph{what}, \emph{who} and \emph{how} questions often pertain to recognition tasks with spatial reasoning. The \emph{where}, \emph{when} and \emph{why} questions usually involve high-level common sense reasoning. The second row depicts \emph{pointing} (\emph{which}) questions where the yellow box is the correct answer and the red boxes are human-generated wrong answers. These four answers form a multiple-choice test for each question.}
\label{fig:mc-qualitative-example}
\vspace{-1mm}
\end{figure*}


\subsection{Collecting the 7W Questions}
The data collection tasks are conducted on Amazon Mechanical Turk (AMT), an online crowdsourcing platform. The online workers are asked to write pairs of question and answer based on image content.
We instruct the workers to be concise and unambiguous to avoid wordy or speculative questions.
To obtain a clean set of high-quality QA pairs, we ask three AMT workers to label each pair as \emph{good} or \emph{bad} independently. The workers judge each pair by whether an average person is able to tell the answer when seeing the image. We accept the QA pairs with at least two positive votes.
We notice varying acceptance rates between categories, ranging from 92\% for \emph{what} to 63\% for \emph{why}. The overall acceptance rate is 85.8\%.

VQA~\cite{antol2015vqa} relied on both human workers and automatic methods to generate a pool of candidate  answers. 
We find that human-generated answers produce the best quality; on the contrary, automatic methods are prone to introducing candidate answers paraphrasing the ground-truth answers. 
For the \emph{telling} questions, the human workers write three plausible answers to each question without seeing the image. To ensure the uniqueness of correct answers, we provide the ground-truth answers to the workers, and instruct them to write answers of different meanings. For the \emph{pointing} questions, the workers draw three bounding boxes of other objects in the image, ensuring that these boxes cannot be taken as the correct answer.  We provide examples from the 7W categories in Fig.~\ref{fig:mc-qualitative-example}.


\begin{figure}[thb]
\begin{center}
\includegraphics[width=1.\linewidth]{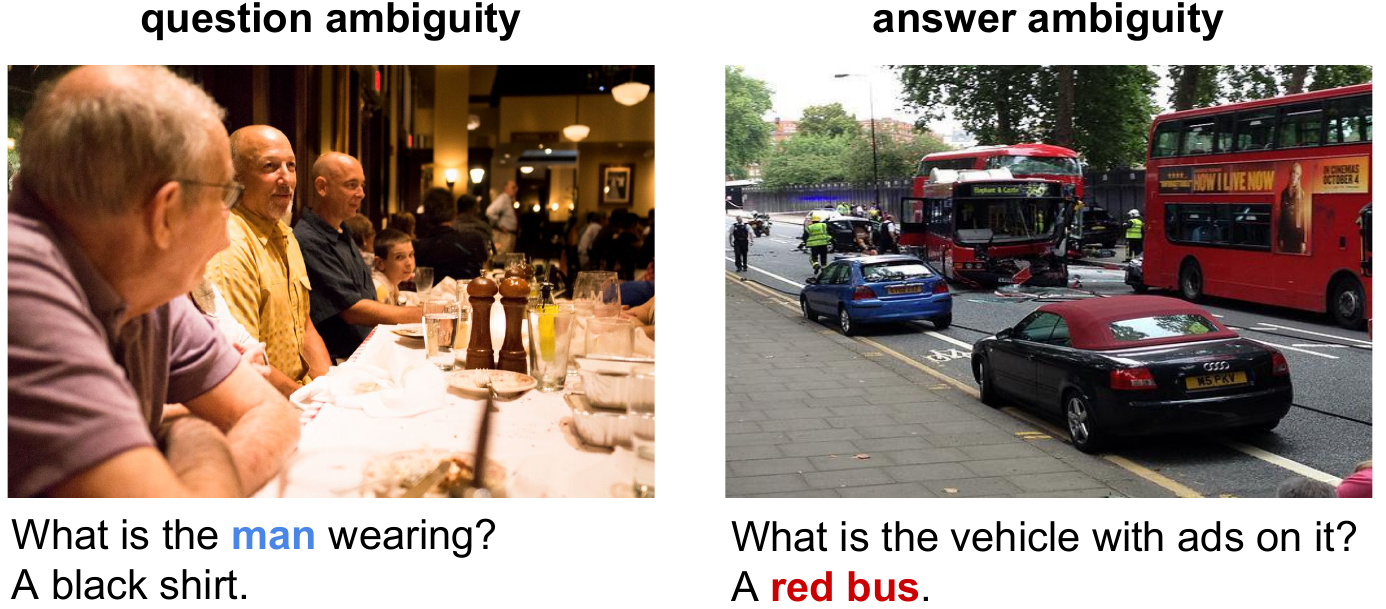}
\vspace{-6mm}
\caption{Coreference ambiguity arises when an object mention has multiple correspondences in an image, and the textual context is insufficient to tell it apart.
The answer to the left question can be either \emph{gray}, \emph{yellow} or \emph{black}, depending on which man is meant.
In the right example, the generic phrase \emph{red bus} can refer to both buses in the image. Thus an algorithm might answer correctly even if referring to the wrong bus.}
\label{fig:6w_qa_ambiguity}
\vspace{-0.25 in}
\end{center}
\end{figure} 


\begin{table*}[t!]
\centering
\caption{Comparisons on Existing Visual Question Answering Datasets}
\vspace{-3mm}
\begin{scriptsize}
\begin{tabular}{lrrccccccccc}
\hline
& \textbf{\# QA} & \textbf{\# Images} & \textbf{AvgQLen} & \textbf{AvgALen} & \textbf{LongAns} & \textbf{TopAns} & \textbf{HumanPerf} & \textbf{COCO} & \textbf{MC} & \textbf{Grounding} & \textbf{VisualAns} \\
\hline
\hline
\textbf{DAQUAR}~\cite{malinowski2014multi} & 12,468 & 1,447 & 11.5$\,\pm\,$2.4 & 1.2$\,\pm\,$0.5 & 3.4\% & 96.4\% &  &  &  &  &  \\
\textbf{Visual Madlibs}~\cite{VisualMadlibs} & 56,468 & 9,688 & 4.9$\,\pm\,$2.4 & 2.8$\,\pm\,$2.0 & 47.4\% & 57.9\% &  &  & $\checkmark$ &  & \\
\textbf{COCO-QA}~\cite{ren2015image} & 117,684 & 69,172 & 8.7$\,\pm\,$2.7 & 1.0$\,\pm\,$0 & 0.0\% & 100\% &  & $\checkmark$ &  &  &  \\
\textbf{Baidu~\cite{gao2015you}} & 316,193 & 316,193 & - & - & - & - & & $\checkmark$ &  &  &  \\
\textbf{VQA}~\cite{antol2015vqa} & 614,163 & 204,721 & 6.2$\,\pm\,$2.0 & 1.1$\,\pm\,$0.4 & 3.8\% & 82.7\% & $\checkmark$ & $\checkmark$ & $\checkmark$ &  &  \\
\textbf{Visual7W (Ours)} & 327,939 & 47,300 & 6.9$\,\pm\,$2.4 & 2.0$\,\pm\,$1.4 & 27.6\% & 63.5\% & $\checkmark$ & $\checkmark$ & $\checkmark$ & $\checkmark$ & $\checkmark$\\
\hline
\end{tabular}
\label{table:comparisons_between_datasets}
\end{scriptsize}
\vspace{-1mm}
\end{table*}


\subsection{Collecting Object-level Groundings}
\label{sec:grounding_canonicalization}

We collect object-level groundings by linking the object mentions in the QA pairs to their bounding boxes in the images. We ask the AMT workers to extract the object mentions from the QA pairs and draw boxes on the images. We collect additional groundings for the multiple choices of the \emph{pointing} questions. Duplicate boxes are removed based on the object names with an Intersection-over-Union threshold of 0.5. In total, we have collected 561,459 object bounding boxes, on average 12 boxes per image.

The benefits of object-level groundings are three-fold: 1) it resolves the coreference ambiguity problem between QA sentences and images; 2) it extends the existing visual QA setups to accommodate visual answers; and 3) it offers a means to understand the distribution of objects, shedding light on the essential knowledge to be acquired for tackling the QA tasks (see Sec.~\ref{sec:data_analysis}).

We illustrate examples of coreference ambiguity in Fig.~\ref{fig:6w_qa_ambiguity}. Ambiguity might cause a question to have more than one plausible answers at test time, thus complicating evaluation.
Our online study shows that, such ambiguity occurs in 1\% of the accepted questions and 7\% of the accepted answers.
This illustrates a drawback of existing visual QA setups~\cite{antol2015vqa,gao2015you,malinowski2014multi,ren2015image,VisualMadlibs}, where in the absence of object-level groundings the textual questions and answers are only loosely coupled to the images.


\section{Comparison and Analysis}
\label{sec:data_analysis}

In this section, we analyze our Visual7W dataset  collected on COCO images (cf. Table~\ref{table:comparisons_between_datasets}, \emph{COCO}), present its key features, and provide comparisons of our dataset with previous work. We summarize important metrics of existing visual QA datasets in Table~\ref{table:comparisons_between_datasets}.\footnote{We report the statistics of VQA dataset~\cite{antol2015vqa} with its real images and Visual Madlibs~\cite{VisualMadlibs} with its filtered hard tasks. 
The fill-in-the-blank tasks in Visual Madlibs~\cite{VisualMadlibs}, where the answers are sentence fragments, differ from other QA tasks, resulting in distinct statistics.
We omit some statistics for Baidu~\cite{gao2015you} due to its partial release.}

\begin{figure}[t]
\begin{center}
\includegraphics[width=1.\linewidth]{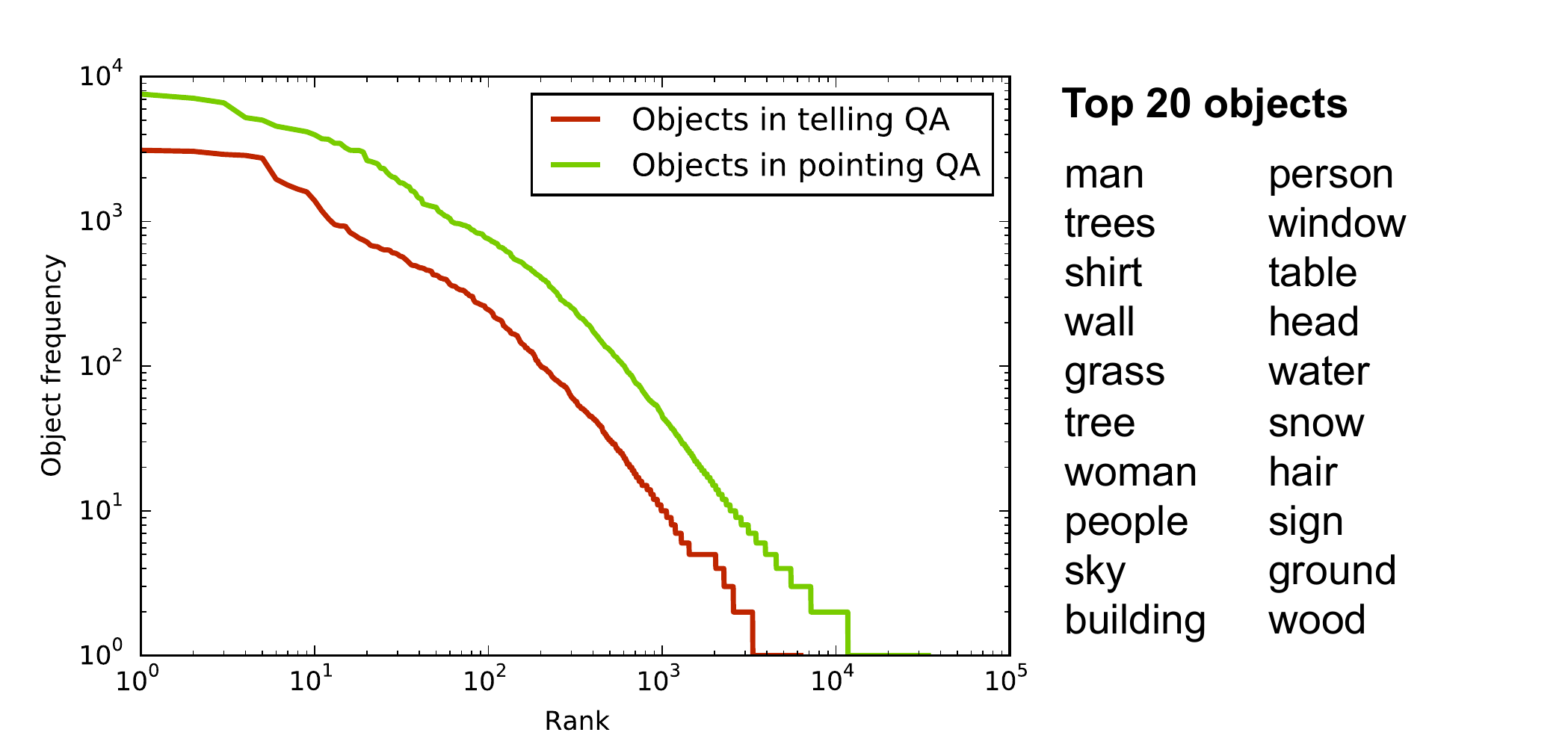}
\vspace{-7mm}
\caption{Object distribution in \emph{telling} and \emph{pointing} QA. The rank of an object category is based on its frequency with rank \#1 referring to the most frequent one. The \emph{pointing} QA pairs cover an order of magnitude more objects than the \emph{telling} QA pairs. The top 20 object categories indicate that the object distribution's bias towards persons, daily-life objects and natural entities.}
\label{fig:object_grounding_distributions}
\vspace{-3mm}
\end{center}
\end{figure}

\vspace{1mm}
\noindent
\textbf{Advantages of Grounding}\quad The unique feature of our Visual7W dataset is the grounding annotations of all textually mentioned objects (cf. Table~\ref{table:comparisons_between_datasets}, \emph{Grounding}). In total we have collected 561,459 object groundings, which enables the new type of visual answers  in the form of bounding boxes (cf. Table~\ref{table:comparisons_between_datasets}, \emph{VisualAns}).
Examining the object distribution in the QA pairs sheds light on the focus of the questions and the essential knowledge to be acquired for answering them. Our object groundings spread across 36,579 categories (distinct object names), thereby exhibiting a long tail pattern where 85\% of the categories have fewer than 5 instances (see Fig.~\ref{fig:object_grounding_distributions}). The open-vocabulary annotations of objects, in contrast with traditional image datasets focusing on predefined categories and salient objects~\cite{lin2014microsoft,ILSVRC15}, provide a broad coverage of objects in the images.

\vspace{1mm}
\noindent
\textbf{Human-Machine Performance Gap}\quad We expect that a good QA benchmark should exhibit a sufficient performance gap between humans and state-of-the-art models, leaving room for future research to explore. Additionally a nearly perfect human performance is desired to certify the quality of its questions. On Visual7W, we conducted two experiments to measure human performance (cf. Table~\ref{table:comparisons_between_datasets}, \emph{HumanPerf}), as well as examining the percentage of questions that can be answered without images. Our results show both strong human performance and a strong interdependency between images and QA pairs. We provide the detailed analysis and comparisons with the state-of-the-art automatic models in Sec.~\ref{sec:exp}.


\begin{table}[htb]
\caption{Model and Human Performances on QA Datasets}
\vspace{-6mm}
\begin{footnotesize}
\begin{center}
\begin{tabular}{|lccc|}
\hline
 & \textbf{Model} & \textbf{Human} & $\Delta$\\
 \hline
 \hline
\textbf{VQA} (open-ended)~\cite{antol2015vqa} & 0.54 & 0.83 & 0.29\\
\textbf{VQA} (multiple-choice)~\cite{antol2015vqa} & 0.57 & 0.92 & 0.35\\
\textbf{Facebook bAbI}~\cite{weston2015towards} & 0.92 & $\sim$1.0 & 0.08\\
\textbf{Ours} (\emph{telling} QA) & 0.54 & 0.96 & \textbf{0.42}\\
\textbf{Ours} (\emph{pointing} QA) & 0.56 & 0.97 & 0.41\\
\hline
\end{tabular}
\end{center}
\end{footnotesize}
\label{tbl:hum-com-gap}
\vspace{-0.17 in}
\end{table}%


Table~\ref{tbl:hum-com-gap} compares Visual7W with VQA~\cite{antol2015vqa} and Facebook bAbI~\cite{weston2015towards}, which have reported model and human performances. Facebook bAbI~\cite{weston2015towards} is a textual QA dataset claiming that humans can potentially achieve 100\% accuracy yet without explicit experimental proof. For VQA~\cite{antol2015vqa}, numbers are reported for both multiple-choice and open-ended evaluation setups. Visual7W features the largest performance gap ($\Delta$), a desirable property for a challenging and long-lasting evaluation task. At the same time, the nearly perfect human performance proves high quality of the 7W questions.

\vspace{1mm}
\noindent
\textbf{QA Diversity}\quad The diversity of QA pairs is an important feature of a good QA dataset as it reflects a broad coverage of image details, introduces complexity and potentially requires a broad range of skills for solving the questions. To obtain diverse QA pairs, we decided to rule out binary questions, contrasting Geman et al.'s proposal~\cite{geman2015visual} and VQA's approach~\cite{antol2015vqa}. We hypothesize that this encourages workers to write more complex questions and also prevents inflating answer baselines with simple yes/no answers.

When examining the richness of QA pairs, the length of questions and answers (cf. Table~\ref{table:comparisons_between_datasets}, \emph{AvgQLen}, \emph{AvgALen}) is a rough indicator for the amount of information and complexity they contain.
The overall average question and answer lengths are 6.9 and 2.0 words respectively. The \emph{pointing} questions have the longest average question length.
The \emph{telling} questions exhibit a long-tail distribution where 51.2\%, 21.2\%, and 16.6\% of their answers have one, two or three words respectively. Many answers to \emph{where} and \emph{why} questions are phrases and sentences, with an average of 3 words. In general, our dataset features long answers where 27.6\% of the questions have answers of more than two words (cf. Table~\ref{table:comparisons_between_datasets}, \emph{LongAns}). In contrast, 89\% of answers in VQA~\cite{antol2015vqa}, 90\% of answers in DAQUAR~\cite{malinowski2014multi} and all answers in COCO-QA~\cite{ren2015image} are a single word. We also capture more long-tail answers as our 1,000 most frequent answers only account for 63.5\% of all our answers (cf. Table~\ref{table:comparisons_between_datasets}, \emph{TopAns}). Finally we provide human created multiple-choices for evaluation (cf. Table~\ref{table:comparisons_between_datasets}, \emph{MC}).


\section{Attention-based Model for Grounded QA}
\label{sec:attention_model}

The visual QA tasks are visually grounded, as local image regions are pertinent to answering questions in many cases.
For instance, in the first \emph{pointing} QA example of Fig.~\ref{fig:mc-qualitative-example} the regions of the window and the pillows reveal the answer, while other regions are irrelevant to the question. We capture this intuition by introducing a spatial attention mechanism similar to the model for image captioning~\cite{xu2015icml}.

\begin{figure}[t]
\centering
\includegraphics[width=.8\linewidth]{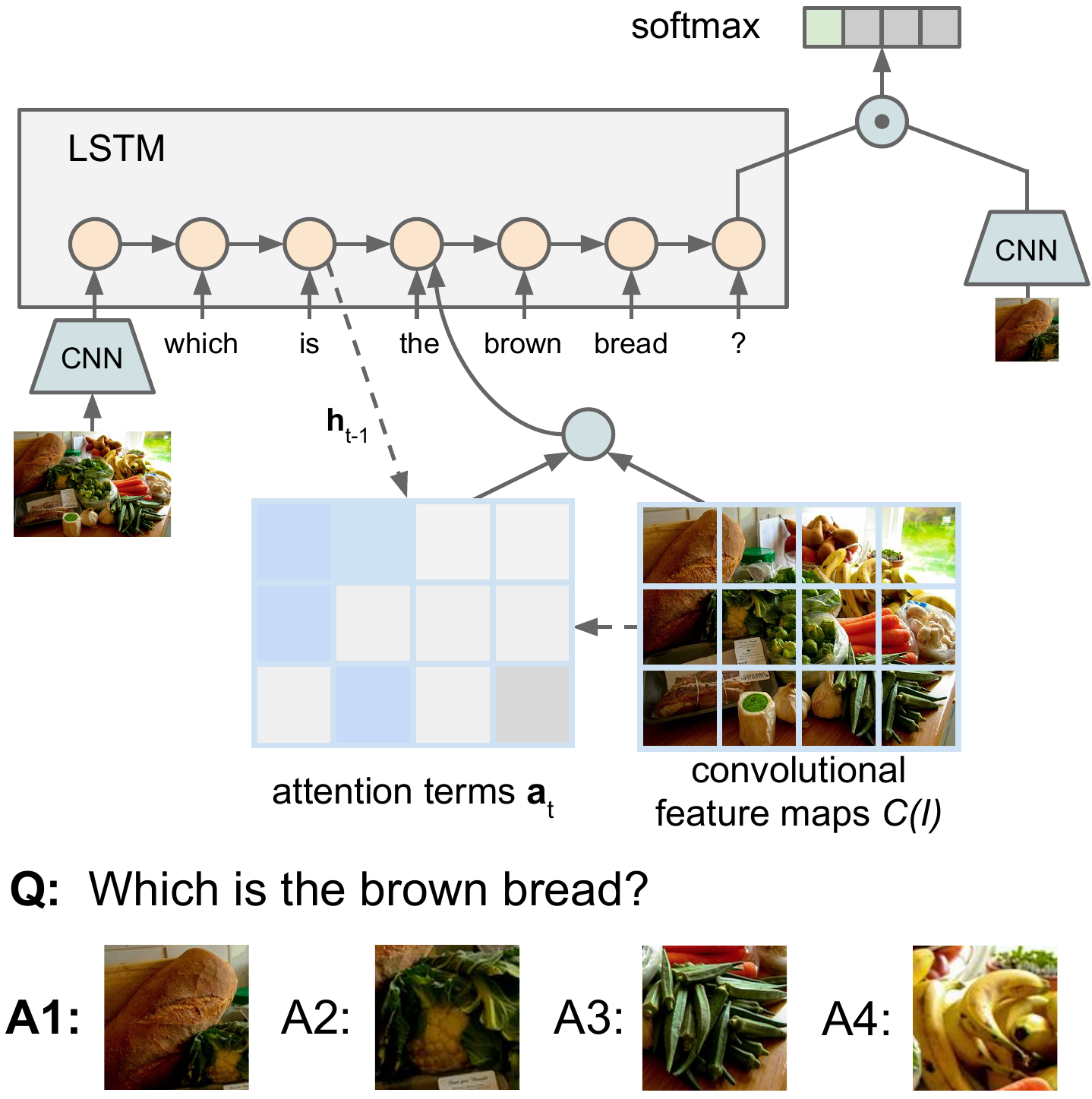}
\vspace{-2mm}
\caption{Diagram of the recurrent neural network model for \emph{pointing} QA. At the encoding stage, the model reads the image and the question tokens word by word. At each word, it computes attention terms based on the previous hidden state and the convolutional feature map, deciding which regions to focus on. At the decoding stage, it computes the log-likelihood of an answer by a dot product between its transformed visual feature (fc7) and the last LSTM hidden state.}
\label{fig:visual6w-lstm-diagram}
\vspace{-2mm}
\end{figure}

\subsection{Recurrent QA Models with Spatial Attention}
LSTM models~\cite{hochreiter1997long} have achieved state-of-the-art results in several sequence processing tasks~\cite{Donahue_2015_CVPR,karpathy2015cvpr,sutskever2014sequence}. 
They have also been used to tackle visual QA tasks~\cite{antol2015vqa,gao2015you,malinowski2015ask}. 
These models represent images by their global features, lacking a mechanism to understand local image regions. 
We add spatial attention~\cite{gregor2015draw,xu2015icml} to the standard LSTM model for visual QA, illustrated in Fig.~\ref{fig:visual6w-lstm-diagram}.
We consider QA as a two-stage process~\cite{gao2015you,malinowski2015ask}. At the encoding stage, the model  memorizes the image and the question into a hidden state vector (the gray box in Fig.~\ref{fig:visual6w-lstm-diagram}). At the decoding stage, the model selects an answer from the multiple choices based on its memory (the \emph{softmax} layer in Fig.~\ref{fig:visual6w-lstm-diagram}). We use the same encoder structure for all visual QA tasks but different decoders for the \emph{telling} and \emph{pointing} QA tasks. Given an image $I$ and a question $Q=(q_1, q_2, \ldots, q_m)$, we learn the embeddings of the image and the word tokens as follow:
\begin{eqnarray}
\small
v_0 & = & W_i[F(I)] + b_i\\
v_i  & = & W_w[OH(t_i)], i = 1, \ldots, m
\end{eqnarray}
where $F(\cdot)$ transforms an image $I$ from pixel space to a 4096-dimensional feature representation. We extract the activations from the last fully connected layer (fc7) of a pre-trained CNN model VGG-16~\cite{simonyan2014very}. $OH(\cdot)$ transforms a word token to its one-hot representation, an indicator column vector where there is a single one at the index of the token in the word vocabulary. The $W_i$ matrix transforms the 4096-dimensional image features into the $d_i$-dimensional embedding space $v_0$, and the $W_w$ transforms the one-hot vectors into the $d_w$-dimensional embedding space $v_i$. We set $d_i$ and $d_w$ to the same value of 512. Thus, we take the image as the first input token.
These embedding vectors $v_{0, 1, \ldots, m}$ are fed into the LSTM model one by one. The update rules of our LSTM model can be defined as follow:
\begin{eqnarray}
\small
\mathbf{i}_t & = & \sigma(W_{vi}v_t + W_{hi}\mathbf{h}_{t-1}+W_{ri}\mathbf{r}_{t}+b_i) \\
\mathbf{f}_t & = & \sigma(W_{vf}v_t + W_{hf}\mathbf{h}_{t-1}+W_{rf}\mathbf{r}_{t}+b_f) \\
\mathbf{o}_t & = & \sigma(W_{vo}v_t + W_{ho}\mathbf{h}_{t-1}+W_{ro}\mathbf{r}_{t}+b_o) \\
\mathbf{g}_t & = & \phi(W_{vg}v_t + W_{hg}\mathbf{h}_{t-1}+W_{rg}\mathbf{r}_{t}+b_g) \\
\mathbf{c}_t & = & \mathbf{f}_t \odot \mathbf{c}_{t-1} + \mathbf{i}_t\odot \mathbf{g}_t\\
\mathbf{h}_t & = & \mathbf{o}_t \odot \phi (\mathbf{c}_t)
\end{eqnarray}
where $\sigma(\cdot)$ is the sigmoid function, $\phi(\cdot)$ is the tanh function, and $\odot$ is the element-wise multiplication operator. The attention mechanism is introduced by the term $\mathbf{r}_{t}$, which is a weighted average of convolutional features that depends upon the previous hidden state and the convolutional features. The exact formulation is as follows:
\begin{eqnarray}
\small
\mathbf{e}_{t}  & = &   w_a^T\tanh({W_{he}\mathbf{h}_{t-1} + W_{ce}C(I)})+b_a \\
\label{eq:attention_terms}
\mathbf{a}_{t}  & = &  \text{softmax}(\mathbf{e}_{t})\\
\mathbf{r}_{t} & = & \mathbf{a}_t^TC(I)
\end{eqnarray}
where $C(I)$ returns the $14\times14$ 512-dimensional convolutional feature maps of image $I$  from the fourth convolutional layer from the same VGG-16 model~\cite{simonyan2014very}. The attention term $\mathbf{a}_{t}$ is a 196-dimensional unit vector, deciding the contribution of each convolutional feature at the $t$-th step. The standard LSTM model can be considered as a special case with each element in $\mathbf{a}_{t}$ set uniformly. $W_i$, $b_i$, $W_w$ and all the $W$s and $b$s in the LSTM model and attention terms are learnable parameters.

\subsection{Learning and Inference}
The model first reads the image $v_0$ and all the question tokens $v_{q_1}$, $v_{q_2}, \ldots, v_{q_m}$ until reaching the question mark (i.e., end token of the question sequence). 
When training for \emph{telling} QA, we continue to feed the ground-truth answer tokens $v_{a_1}$, $v_{a_2}, \ldots, v_{a_n}$ into the model.
For \emph{pointing} QA, we compute the log-likelihood of an candidate region by a dot product between its transformed visual feature (fc7) and the last LSTM hidden state (see Fig.~\ref{fig:visual6w-lstm-diagram}). We use cross-entropy loss to train the model parameters with backpropagation. 
During testing, we select the candidate answer with the largest log-likelihood. We set the hyperparameters using the validation set. The dimensions of the LSTM gates and memory cells are 512 in all the experiments. The model is trained with Adam update rule~\cite{kingma2014adam}, mini-batch size 128, and a global learning rate of $10^{-4}$.


\section{Experiments}
\label{sec:exp}

We evaluate the human and model performances on the QA tasks. We report a reasonably challenging performance delta leaving sufficient room for future research to explore.

\begin{table*}[t!]
\begin{center}
\caption{Human and model performances in the multiple-choice 7W QA tasks (in accuracy)}
\vspace{-3mm}
\begin{small}
\begin{tabular}{|l|c|c|c|c|c|c|c|c|}
\hline
\multirow{ 2}{*}{\textbf{Method}} & \multicolumn{6}{|c|}{\textbf{Telling}} & \textbf{Pointing} & \multirow{2}{*}{\textbf{Overall}}\\
\cline{2-8}
 & \multicolumn{1}{c|}{\textbf{What}} & \multicolumn{1}{c|}{\textbf{Where}} & \multicolumn{1}{c|}{\textbf{When}} & \multicolumn{1}{c|}{\textbf{Who}} & \multicolumn{1}{c|}{\textbf{Why}} & \textbf{How} & \textbf{Which} &\\
\hline
\hline
Human (Question) & 0.356 & 0.322 & 0.393 & 0.342 & 0.439 & 0.337 & - & 0.353\\
Human (Question + Image) & 0.965 & 0.957 & 0.944 & 0.965 & 0.927 & 0.942 & 0.973 & 0.966\\
\hline
\hline
Logistic Regression (Question) & 0.420 & 0.375 & 0.666 & 0.510 & 0.354 & 0.458 & 0.354 & 0.383\\
Logistic Regression (Image) & 0.408 & 0.426 & 0.438 & 0.415 & 0.337 & 0.303 & 0.256 & 0.305\\
Logistic Regression (Question + Image) & 0.429 & 0.454 & 0.621 & 0.501 & 0.343 & 0.356 & 0.307 & 0.352\\
LSTM (Question) & 0.430 & 0.414 & 0.693 & 0.538 & 0.491 & 0.484 & - & 0.462\\
LSTM (Image) & 0.422 & 0.497 & 0.660 & 0.523 & 0.475 & 0.468 & 0.299 & 0.359\\
LSTM (Question + Image)~\cite{malinowski2015ask} & 0.489 & 0.544 & 0.713 & 0.581 & 0.513 & \textbf{0.503} & 0.521 & 0.521\\
Ours, LSTM-Att (Question + Image) & \textbf{0.515} & \textbf{0.570} & \textbf{0.750} & \textbf{0.595} & \textbf{0.555} & 0.498 & \textbf{0.561} & \textbf{0.556}\\
\hline
\end{tabular}
\end{small}
\label{table:visual6w-baseline-performance-multiple-choices}
\end{center}
\vspace{-1mm}
\end{table*}%

\begin{figure}[t]
\begin{center}
\includegraphics[width=.85\linewidth]{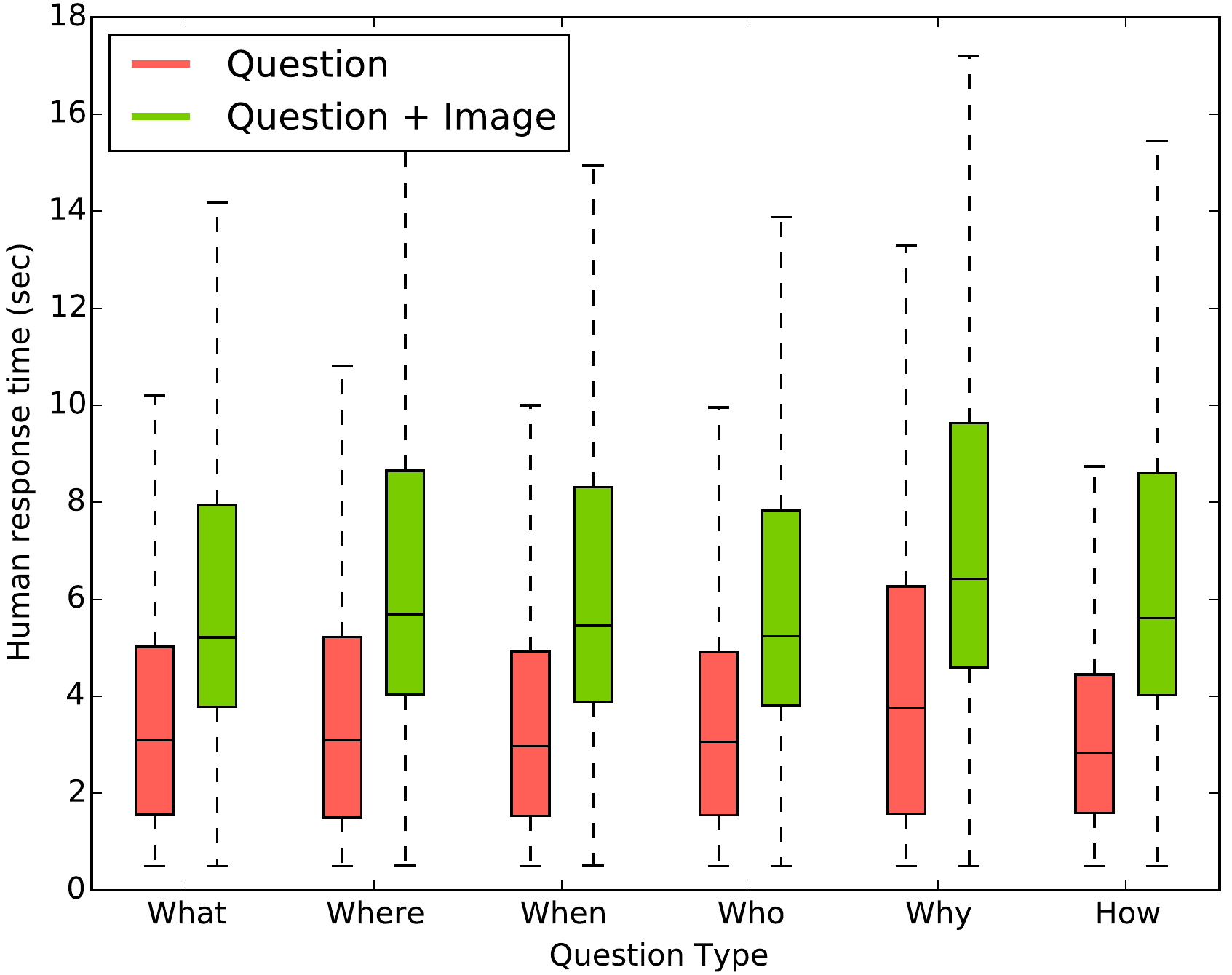}
\vspace{-3mm}
\caption{Response time of human subjects on the \emph{telling} QA tasks. The boxes go from the first quartile to the third quartile of the response time values. The bars in the centers of the boxes indicate the median response time of each category.}
\label{fig:visual6w-human-response-time}
\end{center}
\vspace{-6mm}
\end{figure}

\subsection{Experiment Setups}
\label{sec:exp-setup}
As the 7W QA tasks have been formulated in a multiple-choice format, we use the same procedure to evaluate human and model performances. At test time, the input is an image and a natural language question, followed by four multiple choices. In \emph{telling} QA, the multiple choices are written in natural language; whereas, in \emph{pointing} QA, each multiple choice corresponds to an image region.  We say the model is correct on a question if it picks the correct answer among the candidates. Accuracy is used to measure the performance. 
An alternative method to evaluate \emph{telling} QA is to let the model predict open-ended text outputs~\cite{antol2015vqa}. This approach works well on short answers; however, it performs poorly on long answers, where there are many ways of paraphrasing the same meaning. 
We make the training, validation and test splits, each with 50\%, 20\%, 30\% of the pairs respectively. The numbers are reported on the test set.

\subsection{7W QA Experiments}
\label{sec:visual6w-qa-exps}

\subsubsection{Human Experiments on 7W QA}
\label{sec:visual6w-human-exps}

We evaluate human performances on the multiple-choice 7W QA. We want to measure in these experiments 1) how well humans can perform in the visual QA task and 2) whether humans can use common sense to answer questions without seeing the images.

\begin{figure*}[t!]
\begin{center}
\includegraphics[width=.97\linewidth]{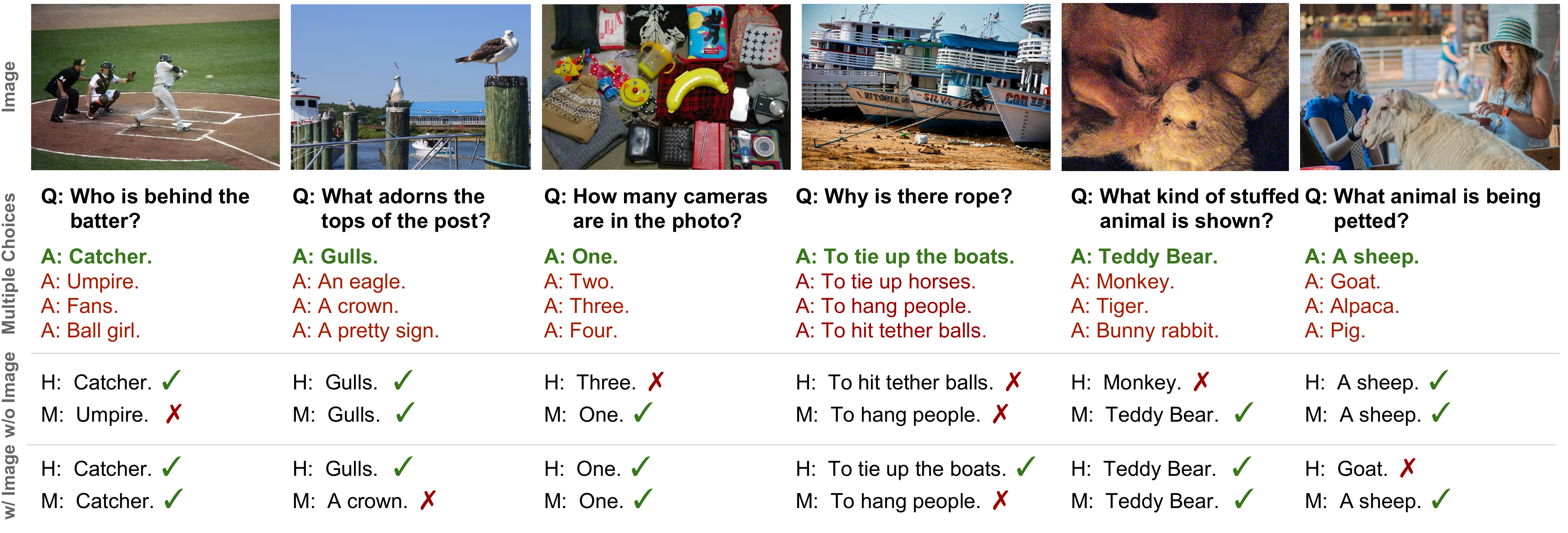}
\vspace{-2mm}
\caption{Qualitative results of human subjects and the state-of-the-art model (LSTM-Att) on multiple-choice QAs. We illustrate the prediction results of six multiple-choice QAs, with and without images. The green answer corresponds to the correct answer to each question, and the rest three are wrong answer candidates. We take the majority votes of five human subjects as the human predictions (H) and the top predictions from the model (M). The correct predictions are indicated by check marks.}
\label{fig:visual6w-qa-prediction-qualitative-results}
\end{center}
\vspace{-5mm}
\end{figure*}

We conduct two sets of human experiments. In the first experiment (Question), a group of five AMT workers are asked to guess the best possible answers from the multiple choices without seeing the images. In the second experiment (Question + Image), we have a different group of five workers to answer the same questions given the images. The first block in Table~\ref{table:visual6w-baseline-performance-multiple-choices} reports the human performances on these experiments. We measure the mean accuracy over the QA pairs where we take the majority votes among the five human responses.
Even without the images, humans manage to guess the most plausible answers in some cases. Human subjects achieve 35.3\% accuracy, 10\% higher than chance. The human performance without images is remarkably high (43.9\%) for the \emph{why} questions, indicating that many \emph{why} questions encode a fair amount of common sense that humans are able to infer without visual cue. However, images are important in the majority of the questions. Human performance is significantly improved when the images are provided. Overall, humans achieve a high accuracy of 96.6\% on the 7W QA tasks. 

Fig.~\ref{fig:visual6w-human-response-time} shows the box plots of response time of the human subjects for \emph{telling} QA. Human subjects spend double the time to respond when the images are displayed. In addition, \emph{why} questions take a longer average response time compared to the other five question types. Human subjects spend an average of 9.3 seconds on \emph{pointing} questions. However, that experiment was conducted in a different user interface, where workers click on the answer boxes in the image. Thus, the response time is not comparable with the \emph{telling} QA tasks.
Interestingly, longer response time does not imply higher performance. 
Human subjects spend more time on questions with lower accuracy. The Pearson correlation coefficient between the average response time and the average accuracy is $-0.135$, indicating a weak negative correlation between the response time and human accuracy.

\subsubsection{Model Experiments on 7W QA}
\label{sec:visual6w-qa-baseline-exps}
Having examined human performance, our next question is how well the state-of-the-art models can perform in the 7W QA task.
We evaluate automatic models on the 7W QA tasks in three sets of experiments: without images (Question), without questions (Image) and with images (Question + Image). In the experiments without images (questions), we zero out the image (questions) features. We briefly describe the three models we used in the experiments:

\vspace{1mm}
\noindent
\textbf{Logistic Regression}\quad A logistic regression model that predicts the answer from a concatenation of image fc7 feature and question feature. The questions are represented by 200-dimensional averaged word embeddings from a pre-trained  model~\cite{mikolov2013distributed}. For \emph{telling} QA, we take the top-5000 most frequent answers (79.2\% of the training set answers) as the class labels. At test time, we select the top-scoring answer candidate. For \emph{pointing} QA, we perform k-means to cluster training set regions by fc7 features into 5000 clusters, used as class labels. At test time, we select the answer candidate closest to the centroid of the predicted cluster.

\vspace{1mm}
\noindent
\textbf{LSTM}\quad The LSTM model in Malinowski and Fritz~\cite{malinowski2015ask} for visual QA with no attention modeling, which can be considered as a simplified version of our full model with the attention terms set to be uniform.

\vspace{1mm}
\noindent
\textbf{LSTM-Att}\quad Our LSTM model with spatial attention introduced in Sec.~\ref{sec:attention_model}, where the attention terms in Eq.~\eqref{eq:attention_terms} determines which region to focus on at each step.

\vspace{1mm}
We report the results in Table~\ref{table:visual6w-baseline-performance-multiple-choices}. All the baseline models surpass the chance performance (25\%).
The logistic regression baseline yields the best performance when only the question features are provided. Having the global image features hurts its performance, indicating the importance of understanding local image regions rather than a holistic representation.
Interestingly, the LSTM performance (46.2\%) significantly outperforms human performance (35.3\%) when the images are not present. Human subjects are not \emph{trained} before answering the questions; however, the LSTM model manages to learn the priors of answers from the training set.
In addition, both the questions and image content contribute to better results.
The Question + Image baseline shows large improvement on overall accuracy (52.1\%) than the ones when either the question or the image is absent.
Finally, our attention-based LSTM model (LSTM-Att) outperforms other baselines on all question types, except the \emph{how} category, achieving the best model performance of 55.6\%.

We show qualitative results of human experiments and the LSTM models on the \emph{telling} QA task in Fig.~\ref{fig:visual6w-qa-prediction-qualitative-results}. 
Human subjects fail to tell a sheep apart from a goat in the last example, whereas the LSTM model gives the correct answer. Yet humans successfully answer the fourth \emph{why} question when seeing the image, where the LSTM model fails in both cases.

\begin{figure}[t]
\begin{center}
\includegraphics[width=0.95\linewidth]{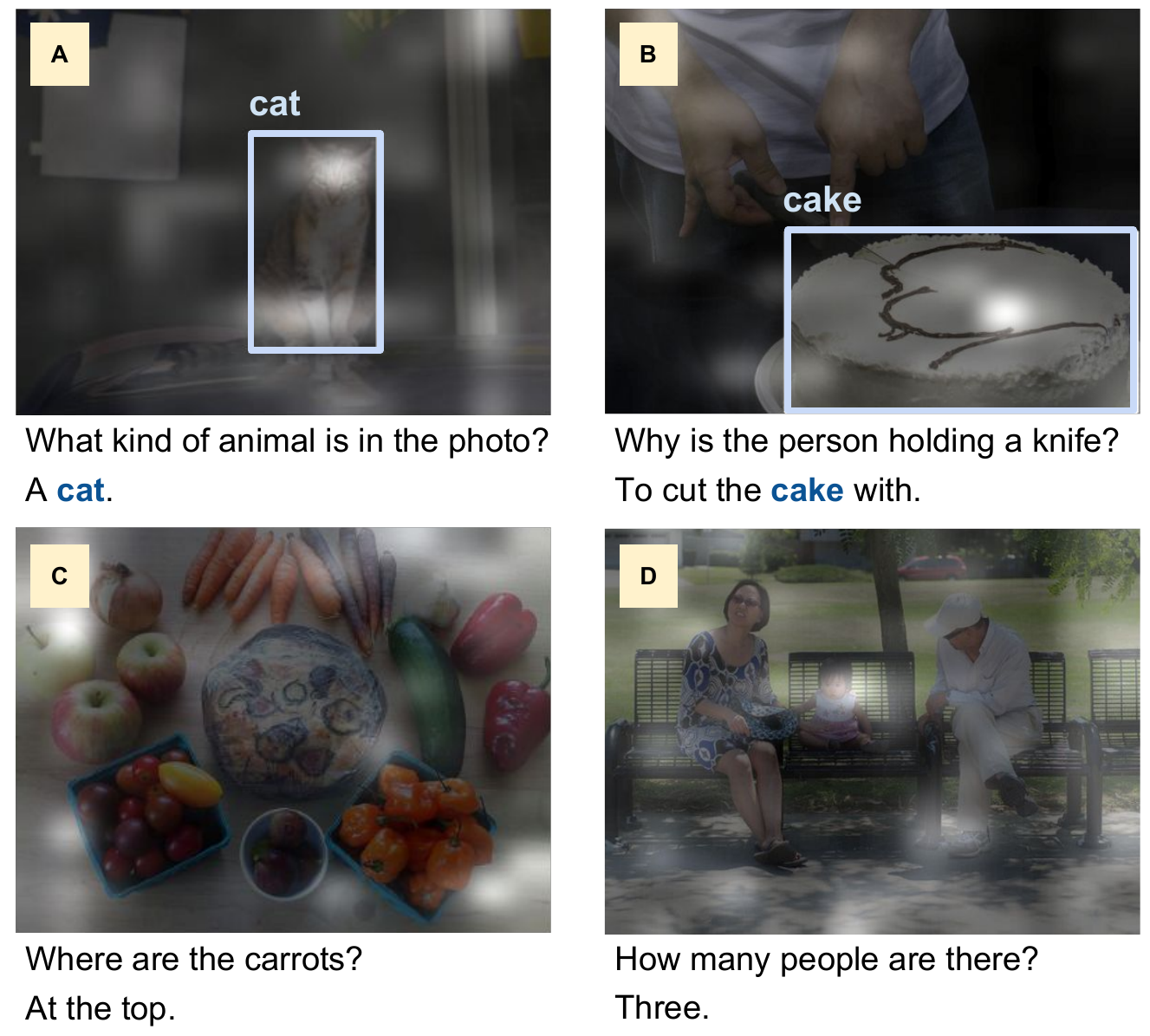}
\vspace{-1mm}
\caption{Object groundings and attention heat maps. We visualize the attention heat maps (with Gaussian blur) on the images. The brighter regions indicate larger attention terms, i.e., where the model focuses. The bounding boxes show the object-level groundings of the objects mentioned in the answers.}
\label{fig:attention-map-visualization}
\end{center}
\vspace{-5mm}
\end{figure}

The object groundings help us analyzing the behavior of the attention-based model. First, we examine where the model focuses by visualizing the attention terms of Eq.~\eqref{eq:attention_terms}. The attention terms vary as the model reads the QA words one by one. We perform max pooling along time to find the maximum attention weight on each of the 14$\times$14 image grid, producing an attention heat map. We see if the model attends to the mentioned objects. The answer object boxes occupy an average of 12\% of image area; while the peak of the attention heat map resides in answer object boxes 24\% of the time. That indicates a tendency for the model to attend to the answer-related regions. We visualize the attention heat maps on some example QA pairs in Fig.~\ref{fig:attention-map-visualization}. The top two examples show QA pairs with answers containing an object. The peaks of the attention heat maps reside in the bounding boxes of the target objects. The bottom two examples show QA pairs with answers containing no object. The attention heat maps are scattered around the image grid. For instance, the model attends to the four corners and the borders of the image to look for the carrots in Fig.~\ref{fig:attention-map-visualization}(c).

\begin{figure}[t]
\begin{center}
\includegraphics[width=.75\linewidth]{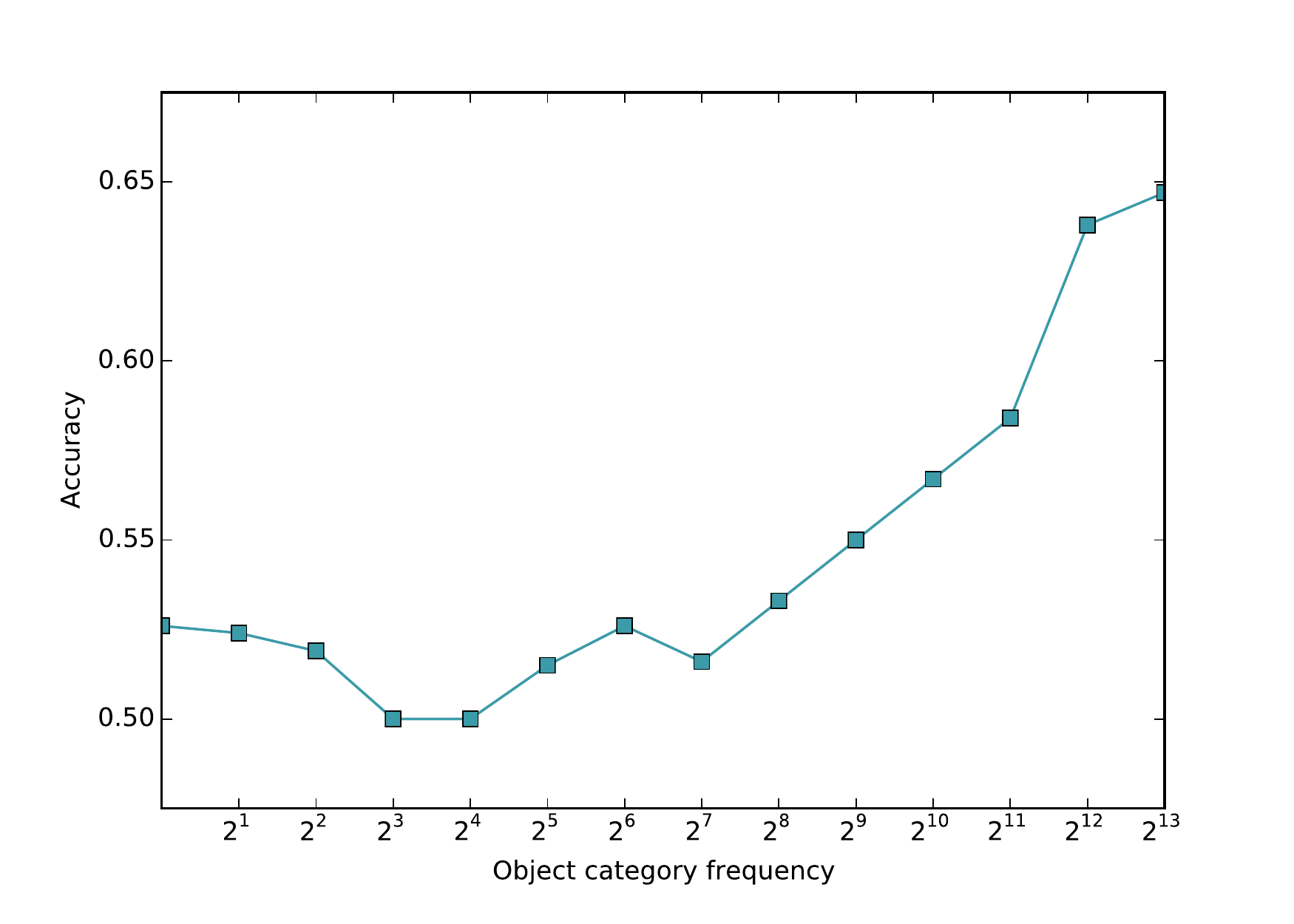}
\vspace{-1mm}
\caption{Impact of object category frequency on the model accuracy in the \emph{pointing} QA task. The $x$-axis shows the upper bound object category frequency of each bin.
The $y$-axis shows the mean accuracy within each bin. The accuracy increases gradually as the model sees more instances from the same category. Meanwhile, the model manages to handle infrequent categories by transferring knowledge from larger categories.}
\label{fig:object-frequncy-on-accuracy}
\end{center}
\vspace{-5mm}
\end{figure}

Furthermore, we use object groundings to examine the model's behavior on the \emph{pointing} QA. Fig.~\ref{fig:object-frequncy-on-accuracy} shows the impact of object category frequency on the QA accuracy. We divide the object categories into different bins based on their frequencies (by power of 2) in the training set. We compute the mean accuracy over the test set QA pairs within each bin. We observe increased accuracy for categories with more object instances. However, the model is able to transfer knowledge from common categories to rare ones, generating an adequate performance (over 50\%) on object categories with only a few instances.


\section{Conclusions}
In this paper, we propose to leverage the visually grounded 7W questions to facilitate a deeper understanding of images beyond recognizing objects. Previous visual QA works lack a tight semantic link between textual descriptions and image regions. We link the object mentions to their bounding boxes in the images. Object grounding allows us to resolve coreference ambiguity, understand object distributions, and evaluate on a new type of visually grounded QA. We propose an attention-based LSTM model to achieve the state-of-the-art performance on the QA tasks. Future research directions include exploring ways of utilizing common sense knowledge to improve the model's performance on QA tasks that require complex reasoning.

\noindent
\textbf{Acknowledgements} We would like to thank Carsten Rother from Dresden University of Technology for establishing the collaboration between the Computer Vision Lab Dresden and the Stanford Vision Lab which enabled Oliver Groth to visit Stanford to contribute to this work. We would also like to thank Olga Russakovsky, Lamberto Ballan, Justin Johnson and anonymous reviewers for useful comments. This research is partially supported by a Yahoo Labs Macro award, and an ONR MURI award.


{\small
\bibliographystyle{ieee}
\bibliography{refs}

\begin{thebibliography}{10}\itemsep=-1pt

\bibitem{antol2015vqa}
S.~Antol, A.~Agrawal, J.~Lu, M.~Mitchell, D.~Batra, C.~L. Zitnick, and
  D.~Parikh.
\newblock {VQA}: Visual question answering.
\newblock {\em ICCV}, 2015.

\bibitem{barnard2003matching}
K.~Barnard, P.~Duygulu, D.~Forsyth, N.~De~Freitas, D.~M. Blei, and M.~I.
  Jordan.
\newblock Matching words and pictures.
\newblock {\em The Journal of Machine Learning Research}, 3:1107--1135, 2003.

\bibitem{bigham2010vizwiz}
J.~P. Bigham, C.~Jayant, H.~Ji, G.~Little, A.~Miller, R.~C. Miller, R.~Miller,
  A.~Tatarowicz, B.~White, S.~White, et~al.
\newblock Vizwiz: nearly real-time answers to visual questions.
\newblock In {\em Proceedings of the 23nd annual ACM symposium on User
  Interface Software and Technology}, 2010.

\bibitem{chen2015cvpr}
X.~Chen and C.~L. Zitnick.
\newblock Mind's eye: A recurrent visual representation for image caption
  generation.
\newblock In {\em CVPR}, 2015.

\bibitem{Donahue_2015_CVPR}
J.~Donahue, L.~Anne~Hendricks, S.~Guadarrama, M.~Rohrbach, S.~Venugopalan,
  K.~Saenko, and T.~Darrell.
\newblock Long-term recurrent convolutional networks for visual recognition and
  description.
\newblock In {\em CVPR}, 2015.

\bibitem{ferrucci2010}
D.~Ferrucci et~al.
\newblock Building {Watson}: An overview of the {DeepQA} project.
\newblock {\em AI Magazine}, 2010.

\bibitem{gao2015you}
H.~Gao, J.~Mao, J.~Zhou, Z.~Huang, L.~Wang, and W.~Xu.
\newblock Are you talking to a machine? dataset and methods for multilingual
  image question answering.
\newblock {\em NIPS}, 2015.

\bibitem{geman2015visual}
D.~Geman, S.~Geman, N.~Hallonquist, and L.~Younes.
\newblock Visual turing test for computer vision systems.
\newblock {\em Proceedings of the National Academy of Sciences},
  112(12):3618--3623, 2015.

\bibitem{girshick2014rich}
R.~Girshick, J.~Donahue, T.~Darrell, and J.~Malik.
\newblock Rich feature hierarchies for accurate object detection and semantic
  segmentation.
\newblock In {\em CVPR}, 2014.

\bibitem{gregor2015draw}
K.~Gregor, I.~Danihelka, A.~Graves, and D.~Wierstra.
\newblock Draw: A recurrent neural network for image generation.
\newblock {\em arXiv preprint arXiv:1502.04623}, 2015.

\bibitem{hochreiter1997long}
S.~Hochreiter and J.~Schmidhuber.
\newblock Long short-term memory.
\newblock {\em Neural computation}, 9(8):1735--1780, 1997.

\bibitem{iyyer2014emnlp}
M.~Iyyer, J.~Boyd-Graber, L.~Claudino, R.~Socher, and H.~{Daum\'e III}.
\newblock A neural network for factoid question answering over paragraphs.
\newblock In {\em Empirical Methods in Natural Language Processing}, 2014.

\bibitem{karpathy2015cvpr}
A.~Karpathy and L.~Fei-Fei.
\newblock Deep visual-semantic alignments for generating image descriptions.
\newblock {\em CVPR}, 2015.

\bibitem{karpathy2014deep}
A.~Karpathy, A.~Joulin, and L.~Fei-Fei.
\newblock Deep fragment embeddings for bidirectional image sentence mapping.
\newblock In {\em NIPS}, pages 1889--1897, 2014.

\bibitem{karpathy2014large}
A.~Karpathy, G.~Toderici, S.~Shetty, T.~Leung, R.~Sukthankar, and L.~Fei-Fei.
\newblock Large-scale video classification with convolutional neural networks.
\newblock In {\em CVPR}, 2014.

\bibitem{kazemzadeh2014emnlp}
S.~Kazemzadeh, V.~Ordonez, M.~Matten, and T.~L. Berg.
\newblock Referit game: Referring to objects in photographs of natural scenes.
\newblock In {\em EMNLP}, 2014.

\bibitem{kiapour2015iccv}
M.~H. Kiapour, X.~Han, S.~Lazebnik, A.~C. Berg, and T.~L. Berg.
\newblock Where to buy it: Matching street clothing photos in online shops.
\newblock {\em ICCV}, 2015.

\bibitem{kingma2014adam}
D.~Kingma and J.~Ba.
\newblock Adam: A method for stochastic optimization.
\newblock {\em arXiv preprint arXiv:1412.6980}, 2014.

\bibitem{kong2014you}
C.~Kong, D.~Lin, M.~Bansal, R.~Urtasun, and S.~Fidler.
\newblock What are you talking about? text-to-image coreference.
\newblock In {\em CVPR}, 2014.

\bibitem{krishnavisualgenome}
R.~Krishna, Y.~Zhu, O.~Groth, J.~Johnson, K.~Hata, J.~Kravitz, S.~Chen,
  Y.~Kalantidis, L.-J. Li, D.~A. Shamma, M.~Bernstein, and L.~Fei-Fei.
\newblock Visual genome: Connecting language and vision using crowdsourced
  dense image annotations.
\newblock In {\em arXiv preprint arxiv:1602.07332}, 2016.

\bibitem{krizhevsky2012imagenet}
A.~Krizhevsky, I.~Sutskever, and G.~E. Hinton.
\newblock Imagenet classification with deep convolutional neural networks.
\newblock In {\em NIPS}, pages 1097--1105, 2012.

\bibitem{kuhn2013political}
R.~Kuhn and E.~Neveu.
\newblock {\em Political journalism: New challenges, new practices}.
\newblock Routledge, 2013.

\bibitem{lampert2009learning}
C.~H. Lampert, H.~Nickisch, and S.~Harmeling.
\newblock Learning to detect unseen object classes by between-class attribute
  transfer.
\newblock In {\em CVPR}, 2009.

\bibitem{linlearning}
T.-Y. Lin, Y.~Cui, S.~Belongie, and J.~Hays.
\newblock Learning deep representations for ground-to-aerial geolocalization.
\newblock In {\em CVPR}, 2015.

\bibitem{lin2014microsoft}
T.-Y. Lin, M.~Maire, S.~Belongie, J.~Hays, P.~Perona, D.~Ramanan,
  P.~Doll{\'a}r, and C.~L. Zitnick.
\newblock Microsoft coco: Common objects in context.
\newblock In {\em ECCV}. 2014.

\bibitem{ma2015cnnQA}
L.~Ma, Z.~Lu, and H.~Li.
\newblock Learning to answer questions from image using convolutional neural
  network.
\newblock {\em arXiv preprint arXiv:1506.00333}, 2015.

\bibitem{malinowski2014multi}
M.~Malinowski and M.~Fritz.
\newblock A multi-world approach to question answering about real-world scenes
  based on uncertain input.
\newblock In {\em NIPS}, pages 1682--1690, 2014.

\bibitem{malinowski2015ask}
M.~Malinowski, M.~Rohrbach, and M.~Fritz.
\newblock Ask your neurons: A neural-based approach to answering questions
  about images.
\newblock {\em ICCV}, 2015.

\bibitem{mikolov2013distributed}
T.~Mikolov, I.~Sutskever, K.~Chen, G.~S. Corrado, and J.~Dean.
\newblock Distributed representations of words and phrases and their
  compositionality.
\newblock In {\em NIPS}, 2013.

\bibitem{palermo2012dating}
F.~Palermo, J.~Hays, and A.~A. Efros.
\newblock Dating historical color images.
\newblock In {\em ECCV}. 2012.

\bibitem{patterson2012sun}
G.~Patterson and J.~Hays.
\newblock Sun attribute database: Discovering, annotating, and recognizing
  scene attributes.
\newblock In {\em CVPR}, 2012.

\bibitem{pickup2014seeing}
L.~C. Pickup, Z.~Pan, D.~Wei, Y.~Shih, C.~Zhang, A.~Zisserman, B.~Scholkopf,
  and W.~T. Freeman.
\newblock Seeing the arrow of time.
\newblock In {\em CVPR}, 2014.

\bibitem{pirsiavash2014inferring}
H.~Pirsiavash, C.~Vondrick, and A.~Torralba.
\newblock Inferring the why in images.
\newblock {\em arXiv preprint arXiv:1406.5472}, 2014.

\bibitem{plummer2015flickr30k}
B.~Plummer, L.~Wang, C.~Cervantes, J.~Caicedo, J.~Hockenmaier, and S.~Lazebnik.
\newblock Flickr30k entities: Collecting region-to-phrase correspondences for
  richer image-to-sentence models.
\newblock {\em ICCV}, 2015.

\bibitem{ramanathan2014linking}
V.~Ramanathan, A.~Joulin, P.~Liang, and L.~Fei-Fei.
\newblock Linking people with "their" names using coreference resolution.
\newblock In {\em ECCV}, 2014.

\bibitem{ren2015image}
M.~Ren, R.~Kiros, and R.~Zemel.
\newblock Exploring models and data for image question answering.
\newblock {\em NIPS}, 2015.

\bibitem{rohrbach2013translating}
M.~Rohrbach, W.~Qiu, I.~Titov, S.~Thater, M.~Pinkal, and B.~Schiele.
\newblock Translating video content to natural language descriptions.
\newblock In {\em ICCV}, 2013.

\bibitem{ILSVRC15}
O.~Russakovsky, J.~Deng, H.~Su, J.~Krause, S.~Satheesh, S.~Ma, Z.~Huang,
  A.~Karpathy, A.~Khosla, M.~Bernstein, A.~C. Berg, and L.~Fei-Fei.
\newblock {ImageNet Large Scale Visual Recognition Challenge}.
\newblock {\em IJCV}, pages 1--42, April 2015.

\bibitem{simonyan2014very}
K.~Simonyan and A.~Zisserman.
\newblock Very deep convolutional networks for large-scale image recognition.
\newblock {\em ICLR}, 2014.

\bibitem{socher2014grounded}
R.~Socher, A.~Karpathy, Q.~V. Le, C.~D. Manning, and A.~Y. Ng.
\newblock Grounded compositional semantics for finding and describing images
  with sentences.
\newblock {\em Transactions of the Association for Computational Linguistics},
  2:207--218, 2014.

\bibitem{sutskever2014sequence}
I.~Sutskever, O.~Vinyals, and Q.~V. Le.
\newblock Sequence to sequence learning with neural networks.
\newblock In {\em NIPS}, pages 3104--3112, 2014.

\bibitem{taigman2014deepface}
Y.~Taigman, M.~Yang, M.~Ranzato, and L.~Wolf.
\newblock Deepface: Closing the gap to human-level performance in face
  verification.
\newblock In {\em CVPR}, 2014.

\bibitem{toshev2014deeppose}
A.~Toshev and C.~Szegedy.
\newblock Deeppose: Human pose estimation via deep neural networks.
\newblock In {\em CVPR}, 2014.

\bibitem{tu2014joint}
K.~Tu, M.~Meng, M.~W. Lee, T.~E. Choe, and S.-C. Zhu.
\newblock Joint video and text parsing for understanding events and answering
  queries.
\newblock In {\em IEEE MultiMedia}, 2014.

\bibitem{Vinyals_2015_CVPR}
O.~Vinyals, A.~Toshev, S.~Bengio, and D.~Erhan.
\newblock Show and tell: A neural image caption generator.
\newblock In {\em CVPR}, 2015.

\bibitem{weston2015towards}
J.~Weston, A.~Bordes, S.~Chopra, and T.~Mikolov.
\newblock Towards ai-complete question answering: a set of prerequisite toy
  tasks.
\newblock {\em arXiv preprint arXiv:1502.05698}, 2015.

\bibitem{xu2015icml}
K.~Xu, J.~Ba, R.~Kiros, K.~Cho, A.~Courville, R.~Salakhutdinov, R.~Zemel, and
  Y.~Bengio.
\newblock Show, attend and tell: Neural image caption generation with visual
  attention.
\newblock In {\em ICML}, 2015.

\bibitem{young2014image}
P.~Young, A.~Lai, M.~Hodosh, and J.~Hockenmaier.
\newblock From image descriptions to visual denotations: New similarity metrics
  for semantic inference over event descriptions.
\newblock {\em Transactions of the Association for Computational Linguistics},
  2:67--78, 2014.

\bibitem{VisualMadlibs}
L.~Yu, E.~Park, A.~C. Berg, and T.~L. Berg.
\newblock {Visual Madlibs: Fill in the blank Image Generation and Question
  Answering}.
\newblock {\em ICCV}, 2015.

\bibitem{zhou2014learning}
B.~Zhou, A.~Lapedriza, J.~Xiao, A.~Torralba, and A.~Oliva.
\newblock Learning deep features for scene recognition using places database.
\newblock In {\em NIPS}, 2014.

\bibitem{zhu2014eccv}
Y.~Zhu, A.~Fathi, and L.~Fei-Fei.
\newblock Reasoning about object affordances in a knowledge base
  representation.
\newblock {\em ECCV}, 2014.

\bibitem{zitnick2013learning}
C.~L. Zitnick, D.~Parikh, and L.~Vanderwende.
\newblock Learning the visual interpretation of sentences.
\newblock In {\em ICCV}, 2013.

\end{thebibliography}
}

\end{document}